\documentclass[lettersize,journal]{IEEEtran}
\usepackage{amsmath,amsfonts}
\usepackage{algorithmic}
\usepackage{algorithm}
\pdfoutput=1
\usepackage{array}
\usepackage[caption=false,font=footnotesize,labelfont=rm,textfont=rm]{subfig}
\usepackage{textcomp}
\usepackage{stfloats}
\usepackage{url}
\usepackage{verbatim}
\usepackage{graphicx}
\usepackage{cite}

\usepackage{multirow}
\usepackage{booktabs}
\usepackage{makecell}
\usepackage{amsmath}
\usepackage{bm}
\begin{document}

\title{Autonomous and Adaptive Role Selection for Multi-robot Collaborative Area Search Based on Deep Reinforcement Learning}

\author{Lina Zhu, Jiyu Cheng, Hao Zhang, Zhichao Cui, Wei Zhang,~\IEEEmembership{Senior Member,~IEEE,} and Yuehu Liu
        % <-this % stops a space
\thanks{Lina Zhu and Yuehu Liu are with the Institute of Artificial             Intelligence and Robotics, Xi’an Jiaotong University, Xi’an,            China, 710049 (e-mail: Zhulina@stu.xjtu.edu.cn; liuyh@xjtu.edu.cn).

        Zhichao Cui is with the School of Electronics and Control Engineering, Chang'an University, Xi'an, China, 710064 (e-mail: cui.zhichao@chd.edu.cn).
        
        Jiyu Cheng, Hao Zhang, and Wei Zhang are with the School of Control Science and Engineering, Shandong University, Shandong, China, 250061 (e-mail: 202114807@mail.sdu.edu.cn; { \{jycheng, davidzhang\}@sdu.edu.cn}).
        
        \textit{(Corresponding author: Jiyu Cheng.)}}}% <-this % stops a space

% \markboth{IEEE TRANSACTIONS ON NEURAL NETWORKS AND LEARNING SYSTEMS}%
% {Zhu \MakeLowercase{\textit{et al.}}: Autonomous and Adaptive Role Selection for Multi-robot Collaborative Area Search Based on Deep Reinforcement Learning}


\maketitle

\begin{abstract}
In the tasks of multi-robot collaborative area search, we propose the unified approach for simultaneous mapping for sensing more targets (exploration) while searching and locating the targets (coverage). Specifically, we implement a hierarchical multi-agent reinforcement learning algorithm to decouple task planning from task execution. The role concept is integrated into the upper-level task planning for role selection, which enables robots to learn the role based on the state status from the upper-view. Besides, an intelligent role switching mechanism enables the role selection module to function between two timesteps, promoting both exploration and coverage interchangeably. Then the primitive policy learns how to plan based on their assigned roles and local observation for sub-task execution. The well-designed experiments show the scalability and generalization of our method compared with state-of-the-art approaches in the scenes with varying complexity and number of robots.  
\end{abstract}


\section{Introduction}
    \IEEEPARstart{M}{ulti-robot} area search remains a fundamental problem in robotics due to its widespread applications such as Mars exploration \cite{nilsson2018toward}, disaster response \cite{liu2016multirobot}, and urban search and rescue \cite{hong2019investigating,shree2021exploiting}. Unlike single-robot systems, multi-robot systems offer significant advantages, leveraging cooperation among the robots for enhanced task efficiency and resilient decision-making, particularly in time-critical tasks. This entails the robots performing collaborative mapping (exploration) while simultaneously searching for the targets (coverage). In the decision-making process, the robots need to 1) sense the environment to gather more information about targets, 2) rescue targets, and 3) coordinate among multi-robots to maximize the number of rescued targets in the explored area within the limited timesteps. Here, our focus is on integrating the role concept in the multi-robot area search problems, which decouples upper-level task planning from the low-level task execution.

    The more intuitive way to address sophisticated tasks is to decompose them as small and simple sub-tasks. Therefore, many researchers divided the area search problem into two distinguishing sub-tasks: exploration and coverage, which are resolved in two distinct phases. In these approaches, the robots first select a sub-area generated by the cellular decomposition algorithm (e.g. Voronoi partition \cite{nair2020mr}) to explore, and then the robot sweeps the entire sub-area along a planned path calculated by the coverage path planning methods \cite{samarakoon2022online} for coverage. However, addressing exploration and coverage tasks separately may lead to sub-optimal solutions and increased computational cost, consequently limiting overall task performance. A unified approach is expected to be studied to maximize resource utilization, improve efficiency, and achieve an optimal solution.

    \begin{figure}[!t]
        \vspace{0.5em}
        \centering
        \includegraphics[scale=0.3]{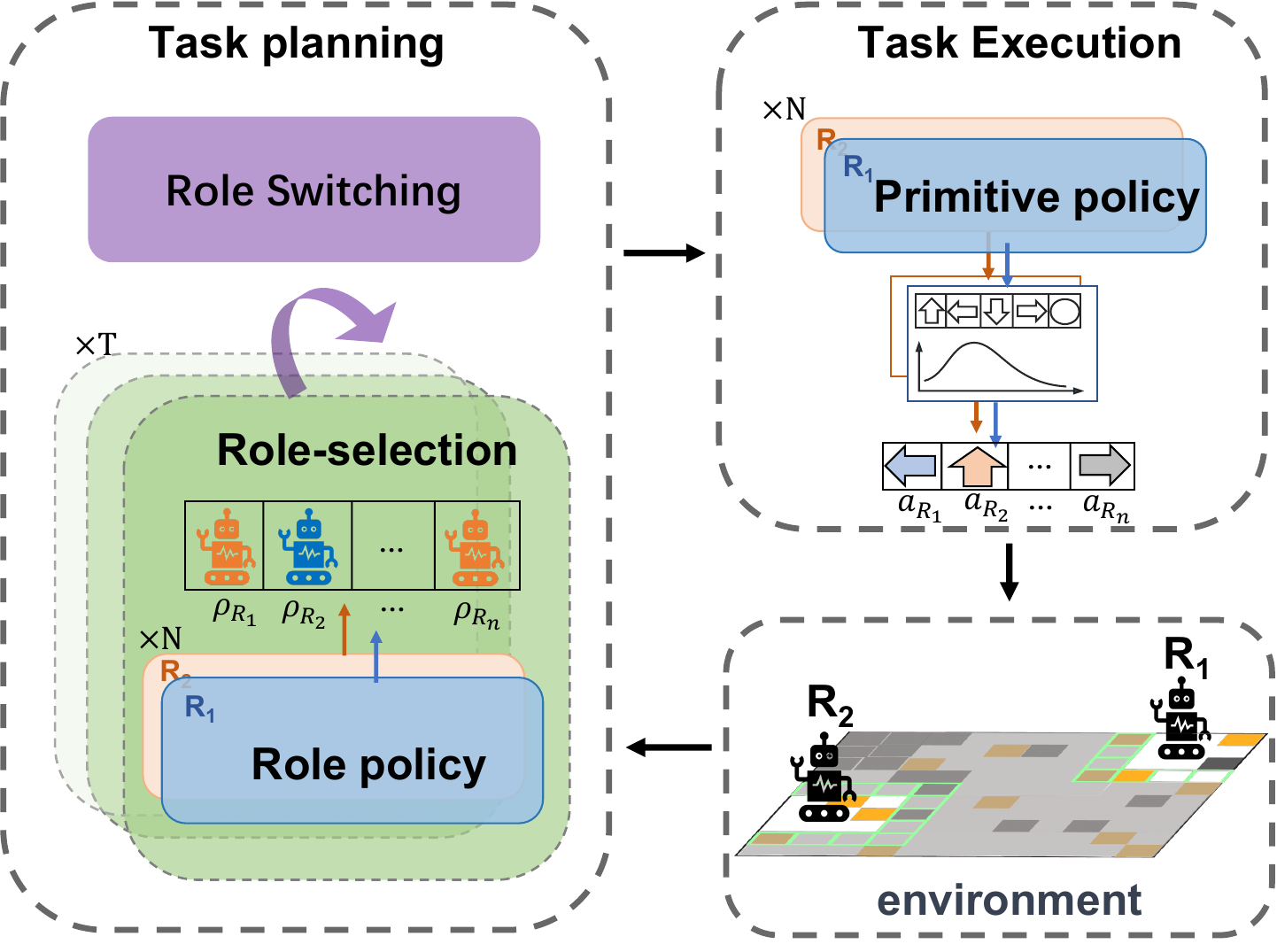}
        \caption{Overview of our method. Our approach involves the separation of training phases for task planning and execution. Task planning utilized a role-selection module, allowing the robot to learn its role from the upper-view. The implementation of role selection between different timesteps is enabled through the role switching mechanism.}
        \label{fig:firstimage}
        \vspace{-0.3cm}
    \end{figure}

    Approaching the aforementioned challenge as a combinatorial optimization problem, the unified approach addresses the  exploration and coverage simultaneously \cite{tolstaya2021multi,zhang2022h2gnn}. They discretized the exploration and coverage problem to induce a graph-represented environment and achieved the simultaneous execution of both sub-tasks by learning the spatial relationship between the sub-tasks and the robots. The above complete unified approaches improve the decision-making capability, but the coupling of task planning and task execution often escalates training complexity. Aiming at reducing the training complexity and realizing explicit cooperation among the robots, decoupling task planning from task execution provides a promising way of integrating upper-level decision-making planners. An advanced comprehension for planning complex tasks necessitates decoupling. Previous solutions like in \cite{patil2023graph} involved decoupling, yet the method for calculating information gain and allocating sub-tasks to each robot is hand-crafted and heuristic. Our approach integrates role concepts into multi-robot coordinated area search problems and trains the upper-level task planning using deep reinforcement learning (DRL), which can scale to high-complexity scenes with various scales and more robots.

    In summary, we present a unified approach that aims for the simultaneous execution of multiple sub-tasks within the multi-robot area search. To overcome the training challenges associated with coupling task planning and task execution, we introduce a role selection module to decouple the task planning from the upper-level view. We train the decentralized role policy using deep reinforcement learning for role selection, guiding robots in choosing between exploring and covering. Additionally, considering that exploration is for coverage and coverage can also be chosen along the planned exploration path, they should complement each other rather than be entirely separated. The intelligent role switching mechanism facilitates mutual reinforcement between these sub-tasks. In guiding the robot to execute corresponding sub-tasks conditioned on the role, we further train a primitive policy with the actor-critic framework.   
    
    Our main contributions are as follows:

    (1) We proposed the role-selection framework for upper-level task planning of the multi-robot area search problem and trained it using multi-agent reinforcement learning, which can guide the multiple robots in selecting their roles to maximize their expertise.
    
    (2) An intelligent role switching mechanism is built to mutually reinforce the roles, which dynamically enhances performance in the sequential role-planning process.
    
    (3) We demonstrate the feasibility of the role-selection framework and training structure by well-designed experiments. We also verify the scalability and generalization of our proposed method against some state-of-the-art methods.

%=======================================================================

\section{Related Work}

\subsection{Multi-Robot Exploration and Coverage}
    In the domain of homogeneous robot systems, researchers have concentrated on various applications within multi-robot exploration and coverage. These applications encompass a wide range of tasks, such as information gathering, active perception, exploration and mapping, region-of-interest reconstruction, and more \cite{tzes2023graph}. Typically, the primary issue in addressing these diverse tasks is either exploration \cite{tian2020search,mcguire2019minimal,zhou2021fuel} or coverage \cite{best2019dec,collins2021scalable,ghassemi2018decentralized}, or a sequential of exploration followed by coverage \cite{10.1145/3306346.3322942,sharma2023d2coplan}, rather than a combination of the both.
    
    To enhance map precision in the larger-scale unknown forest environment for search and rescue, Tian \emph{et al.} \cite{tian2020search} proposed a tree-based map representation method, focusing on lightweight communication during robust loop closure detection. McGuire \emph{et al.} \cite{mcguire2019minimal} introduced a minimal navigation solution tailored for swarms of robots exploring unknown terrains. Zhou \emph{et al.} \cite{zhou2021fuel} devised a frontier information structure for unmanned aerial vehicles (UAVs), enabling fast exploration with incrementally updated. Best \emph{et al.} \cite{best2019dec} developed a variant Monte Carlo tree search (MCTS) method by decentralized planning for multi-robot active perception. In the multi-robot coverage path planning, Collins \emph{et al.} \cite{collins2021scalable} proposed an optimal ordered list of interesting waypoints for each robot, aiming to minimize the costs of coverage path planning and reducing the time of mission completion. 

    Beyond decomposed algorithms, combination methods have emerged to tackle exploration and coverage tasks. These approaches first gather the global knowledge about the environment and then allocate the sub-regions into each robot, calculating the smooth scanning path for individual robots \cite{10.1145/3306346.3322942,sharma2023d2coplan,mitra2022scalable}. Dong \emph{et al.} \cite{10.1145/3306346.3322942} extracted a set of task views for covering scene sub-regions, which are subsequently allocated to the robots. They compute smooth paths for multi-robot collaborative dense reconstruction by each robot. Sharma \emph{et al.} \cite{sharma2023d2coplan} proposed D2CoPlan, addressing multi-robot coverage challenges by guiding decentralized information aggregation and local action selection. Mitra \emph{et al.} \cite{mitra2022scalable} presented a centralized online multi-robot planner, directing robots to explore new regions along the collision paths. These distinct approaches lack uniformity, utilizing different environmental representations and consuming significant computational resources.

\subsection{Role-Based Learning}

    The concept of role, derived from natural human activities, proves to be both prevalent and effective in completing complex collaborative tasks. Consequently, substantial efforts are focused on integrating role division into multi-agent reinforcement learning, involving the decomposition of complex tasks or sub-groups into smaller units for decision-making \cite{bonjean2014adelfe,lhaksmana2018role,zeng2023effective,koley2023opponent}. In role-based frameworks, task complexity is alleviated by decomposing the policy search space into the state-action space of sub-tasks, rather than addressing the total space. However, existing algorithms decomposed have relied on prior knowledge of the state, which is typically unavailable in practice. Wang \emph{et al.} \cite{wang2020roma} proposed the ROMA  over the QMIX \cite{rashid2020monotonic} algorithm, training agents with similar roles to share the parameters and specialize in specific sub-tasks. Additionally, Wang \emph{et al.} \cite{wang2020rode} developed the RODE, a bi-level learning hierarchy that reduces action-observation spaces by decomposing joint action spaces into restricted role action spaces. Liu \emph{et al.} \cite{9859945} introduced the role-oriented methods ROGC, classifying agents into different groups using a role assigner and learning intra-role cooperation based on the graph module. These above-mentioned role-based methods are typically applied to highly challenging cooperative tasks, such as the StarCraft II micromanagement benchmark. In these scenarios, all robots must share a global reward related to joint actions, often neglecting the maximization of local reward. In contrast, our method focuses on training decentralized roles and primitive policies based on local observation. This approach aims to maximize global reward for quick exploration and coverage, while simultaneously focusing on maximizing the local reward to minimize path planning costs.

    \begin{figure*}[!t]
    \centering
    \includegraphics[scale=0.5]{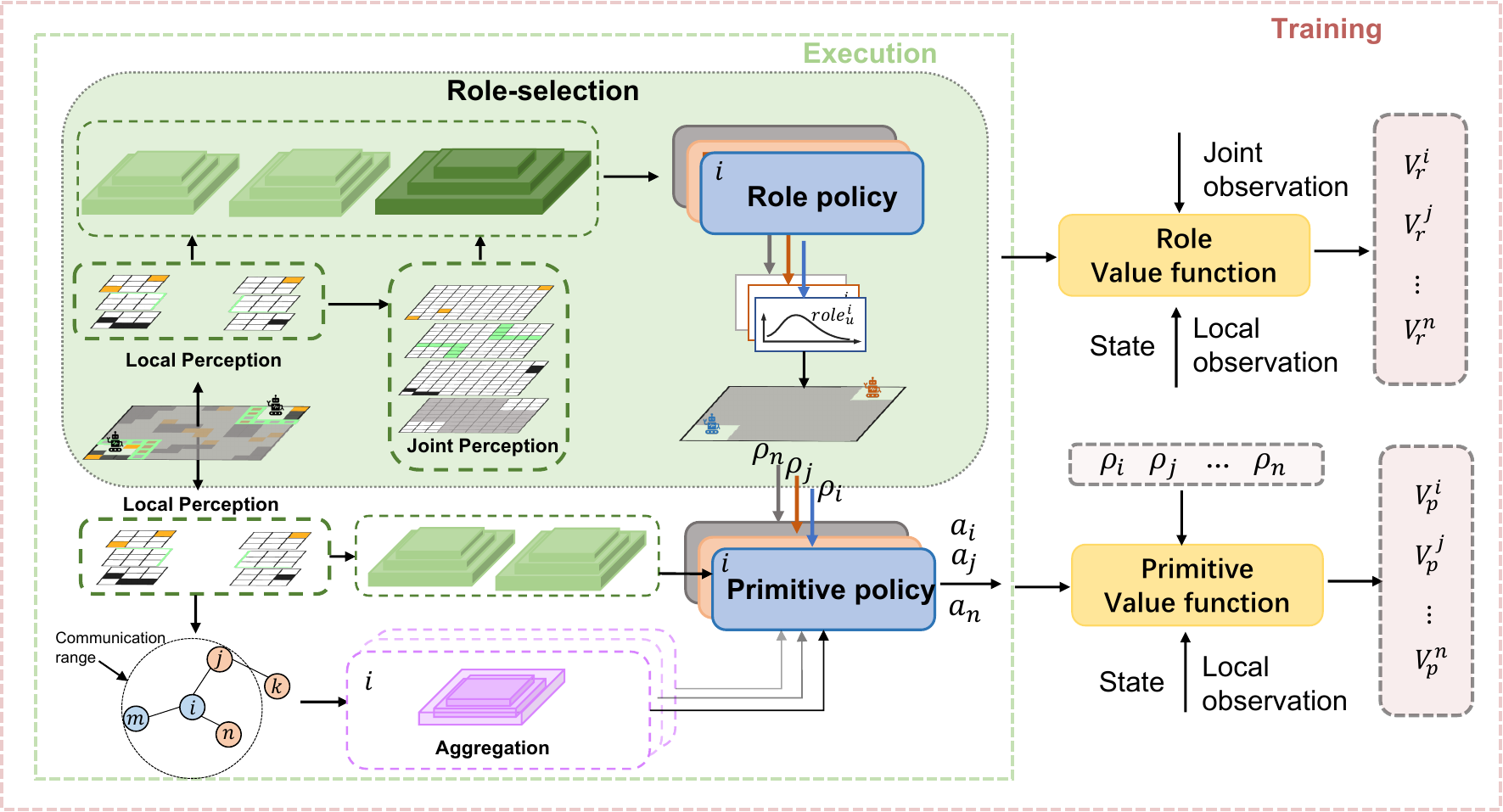}
    \caption{The overall framework of our method. The role-selection module aims to learn the role action of each robot through the role policy from the upper-view. The local and joint observations are encoded as encoding features, which are stacked and used as inputs for the role policy. The role policy generates the role probability distribution for each robot, and the role action is sampled over the role probability distribution. The output results on the global map illustrate the roles. The blue robot at the bottom left represents the \emph{cover} role, while the orange robot at the upper right represents the \emph{explore} role.}
    \label{fig:framework}
    \end{figure*}

%===============================================================

\section{Problem Formulation}

    The environment of the coordinated area search problem is unknown, which is represented as a grid map and consists of two kinds of areas: explored area $|E|$ and unexplored area $|\Bar{E}|$. The whole environment is an unexplored area at the initial timestep. Every robot is equipped with a perception sensor with a fixed sensing range. The cells will be explored if the cells are within the perception range of the robot, which is named after the explored area. The explored area consists of three kinds of cells: obstacles cells, free cells, and target cells. The robot can only move in the free space. 
 
    The tasks are divided into two sub-tasks: exploration and coverage. At each timestep $t$, the robot $i$ receives its egocentric local map and exchanges information of other robots obtained through communication as input and decides on its own role $\rho^i_t$ by the role policy, either to \emph{explore} or to \emph{cover}. Then, the robot outputs its primitive action $a^i_t$ based on the role action $\rho^i_t$ and their local observation for interacting with the environment. Within a short timestep, the goal of the multi-robot area search is to cover as many target cells as possible within the explored area.

    \begin{equation}
        \begin{aligned}
         \pi^* &= \arg\max_{\pi}\sum_{t=0}^T (\sum_{\substack{ i\in N, \\ j\in C_t}} c_t^j \mathbb{I}_{q_t^i=p^j}|\sum_{t=0}^T\sum_{\substack{ i\in N, \\ k\in \bar{E_t}}} u_t^k \mathbb{I}_{q_t^i=p^k}), \\
        s.t.\ 
        c_t^j &= 
        \begin{cases}
            1, & \text{if } q^i_{\tau}=p^j, \forall \tau < t,  j \in C_t,  i\in N \\
            0, & \text{else}
        \end{cases} \\        
         u_t^k &= 
        \begin{cases}
            1,& \text{if } q^i_{\tau}=p^k, \forall \tau < t,  k \in \bar{E_t},  i\in N \\
            0, & \text{else}
        \end{cases}
        \end{aligned}
    \end{equation}

    where $N$ denotes the sets of robots; $C_t$ and $\bar{E_t}$ are the sets of target cells and unexplored cells; $T$ is the maximize training timesteps; the $c_t^j$ is a binary value indicating whether free cell $j$ is the target cell at timestep $t$, $c_t^j =1$ if $j \in C_t$; $\mathbb{I}_{q_t^i=p^j}$ denotes whether robot $i$ arrives at free cell $j$ at timestep $t$. Similarly, $u_t^k$ is a binary value indicating whether free cell $k$ is the unexplored cell, with $u_t^k =1$ if $k \in \bar{E_t}$; $\mathbb{I}_{q_t^i=p^k}$ denotes whether unexplored cell $k$ is within the field of view of robot $i$ at timestep $t$.
    
    % The condition of the objective is related to the explored area, indicating that the target is exclusively detected within its explored area.
    
    % $T$ denotes the sum of the target cells covered by the robot $i$ over the explored area $E$ within a whole episode, and the $E$ denotes the explored area of all robots, $E = E_1 \cup E_i \dots \cup E_n$.
    
%===============================================================================

\section{Goal-augmented Pomdp}

     We formulate the hierarchical policy in the multi-robot area search problem as a decentralized Partially Observable Markov Decision Process enhanced by a goal, named after Goal-Augmented POMDP (GA-POMDP). GA-POMDP is defined by the tuple: $<\mathcal{N},\mathcal{S}, \mathcal{A}, \mathcal{P}, \mathcal{\bar{R}}, \mathcal{\bar{Z}}, \mathcal{\bar{O}}, \gamma>$, where $\mathcal{N}=\{1,2,\dots, N\}$ represents the set of robots. $\mathcal{S}$ denotes the state space of the environment. Every robot $i \in \mathcal{N}$ chooses an action $a^i \in \mathcal{A}^i$, where $\mathcal{A}^i$ denotes the action space of robot $i$. All robots' actions formulate a joint action $\bm{a}=\left\{a^1,a^2,\cdots, a^{N}\right\}$. Let the $\mathcal{A} := \mathcal{A}^1 \times \dots \times \mathcal{A}^N$ be the joint action space. The state transits to any state $s^{\prime} \in \mathcal{S}$ result from the joint action $\bm{a}$ generated by state action probability transition function $\mathrm{P}\left({s}^{\prime}\mid{s}, \bm{a}\right)$: $\mathcal{S} \times \mathcal{A} \to \mathcal{S}^{\prime}$. $\mathcal{\bar{R}}$ is the set of global rewards based on reward function $r(s,\bm{a})$: $\mathcal{S} \times \mathcal{A} \times \mathcal{S}^{\prime} \to \mathbb{R}$, and $\overline{o}^i \in \mathcal{\bar{Z}}^i$ denotes the local observation generated by a global state $s$ and a local observation function $\mathcal{\bar{O}}(s,i)$ : $\mathcal{S} \times \mathcal{N} \to \mathcal{\bar{Z}}$, where $\bar{o}^{i}=\left[o^{i}, g^{i}\right]$ is local observation augmented by the sub-goal $g^{i}$. Each robot has its local action-observation history $\tau^i \in \mathcal{T} \equiv (\mathcal{\bar{Z}}^i \times \mathcal{A}^i)$, and learns policy $\pi^i(a^i|\tau^i)$ : $ \mathcal{T} \times \mathcal{A} \to [0,1]$ based on its local trajectory. The objective of all robots is to maximize the discounted cumulative return $\sum_{t=1}^{T}\gamma^{t}r_t$, where $\gamma \in [0,1]$ is the discount factor.
    
%===============================================================================
\section{Methodology} 
    
    In this section, we introduce a novel double actor-critic MARL that integrates the concept of role into the domain of multi-robot area search tasks. Our approach adopts the CTDE paradigm, a well-established structure in decentralized policy-based multi-robot collaborative tasks \cite{zhang2023nowhere}. Illustrated in Fig. \ref{fig:framework}, our framework employs two sets of actor-critic networks for training, denoted as (Actor-Critic)$^{R}$ and (Actor-Critic)$^{P}$, respectively. The Actor$^R$ is dedicated to role selection, with  its output serving as input to both the Actor$^P$ and Critic$^P$ networks, guiding their functioning. During the execution, the Critic$^P$ and Critic$^R$ are removed from the process, enabling each robot to act based on the role policy established by Actor$^R$ and its local primitive policy governed by Actor$^P$. The robot's action probability distribution of primitive and state value are dependent on the role actions, and each of the roles is responsible for automatically identified sub-tasks. 

\subsection{Learning the Role Policy}

    Our approach introduces task planning for role selection, and then task execution is based on this role action. Moreover, conditioning the role actions of all robots on joint observation signifies an advanced comprehension of the dynamically evolving environments in area search scenarios. This design facilitates adaptability to the high-complexity scenes with various scales and more robots.

    The Multi-agent reinforcement learning (MARL) algorithm is employed to train the role policy using a centralized training and decentralized execution architecture based on an actor-critic structure. The objective is to learn the intentions of robots utilizing a decentralized execution actor network. During centralized training, the critic network generates the state value $V_r(s)$, utilized in calculating the advantage function $A_r(s, \rho)$ to judge the reasonableness of the robots' role output by the Actor$^R$. 
    The subsequent subsections provide a detailed discussion of the core modules comprising the role policy.

\subsubsection{Role Action Space}

    In multi-robot coordinated tasks, our work customizes the number of role action spaces to match the number of sub-tasks. Consequently, we define the role action space with two discrete values corresponding to exploration and coverage sub-tasks: [\emph{explore}, \emph{cover}]. When a robot receives the \emph{explore} action, it is desired to move towards the nearest frontier cell within its field of view (FOV). The detection of frontier cell implementation is based on the frontier-based exploration approach \cite{yamauchi1998frontier}. Conversely, upon receiving the \emph{cover} action, the robot is expected to move toward the nearest target cell.

 \subsubsection{Role Observation Space}
 
    In the area search environment with static obstacles and targets, each robot $i$ operates within FOV, defined by the visible radius $r_{FOV}$ within the FOV. Therefore, each robot $i$ observes a portion of the environment within its FOV. The available local observation $o^i_t=\{o^{o}_t,o^{e}_t,o^{c}_t,o^{p}_t\}$ of robot $i$ at timestep $t$ represented as a 4-channel map with a size of $r_{FOV}*r_{FOV}$, comprising an obstacle map, explored map, covered map, and position map. Specifically, the obstacle map $o^{o}_t$ collects the location of free cells and obstacle cells. Similarly, the explored map $o^{e}_t$ and covered map $o^{c}_t$ identify frontier and target cells, respectively. The position map $o^{p}_t$ collects the location of the neighbor robots $N_i$ that meet the communication condition $\Vert{p_{N_i}}-{p_i}\Vert \leq r_{comm}$. For instance, the robot $j$ located at position $p_j$ is the neighbor of the robot $i$ if $\Vert{p_j}-{p_i}\Vert \leq r_{comm}$.
    
    Significantly, the environment is masked by the unexplored area before we extract the local observation, rendering the area outside the explored area invisible to the robots. We set the unexplored area as $0$ in each channel map using a binarization function.

    In the decision-making process of Actor$^R$, robot $i$ undertakes a dual planning, Firstly, based on the local observation $o^i_t$, the frontier and target cells are identified for exploration and coverage, respectively. Simultaneously, it evaluates the expected return between the exploration and coverage, leveraging joint local observation $jo^i_t$. The joint local observation $jo^i_t=\{jo^{me}_t,jo^{mc}_t\}$ at timestep $t$ consists of the merged explored map $jo^{me}_t$ and the merged covered map $jo^{mc}_t$. Specifically, the merged explored map $jo^{me}_t \in \mathbb{R}^{W\times H}$, $jo^{me}_t = \{o^{e}_0,\dots,o^{e}_{t-1},o^{e}_t\}$ is a union of the historical explored area of all robots, where the $W$ and $H$ denotes the width and height of environment. And the merged covered map $jo^{mc}_t \in \mathbb{R}^{W\times H}$, $jo^{mc}_t=\{o^{c}_0,\dots,o^{c}_{t-1},o^{c}_t\}$ is a union of historical covered map of all robots. Consequently, the combination of local observation and joint local observation constructs the input tensor of the Actor$^R$, which outputs a role probability distribution for the robot.

 \subsubsection{Role Rewards}

    During the training phase, we define two distinct primitive rewards for robot $i$: exploration reward $R_e$ and coverage reward $R_c$. Further elaboration on these rewards is provided in Section V-B. Additionally, we establish the role reward as:
    
    \begin{equation}
        \begin{aligned}
        R_{t}= \alpha R_e + \beta R_c,
        \end{aligned}
    \end{equation}

    where $\alpha$ and $\beta$ are the reward weight coefficients of the \emph{explore} and \emph{cover} role action, respectively. These coefficients serve to modulate the robots' planning based on the environmental complexity. When the $\alpha$ is set as 1 and $\beta$ is set as 0, the robots prioritize exploration, otherwise, they lean towards coverage.   

\subsubsection{Role Policy Optimization}

    For the training of our method, we implement the MAPPO algorithm with the decentralized actors with a centralized critic in both role and primitive policy. Under this paradigm, each robot has a dependent local policy and a centralized value network.

    In the role policy, each robot learns a local role policy $\pi_{\theta_{r}}^{i} (\rho_{t}^{i} \mid o_{t}^{i},jo^i_t): \mathcal{O} \times \mathcal{JO} \to [0,1] $ parameterized by a local set of parameters $\theta_{r}$. The Generalized Advantage Estimate (GAE) is adopted to estimate the advantage of the Actor-Critic structure, which is defined as $\hat{A}_{r}^i = \sum_{t=1}^{\infty}(\gamma \lambda)^{l} \delta^i_r (t+l)$ where $\lambda \in [0,1]$ and TD-error is $\delta^i_r (t) = r^i_t + \lambda V^i_{\phi_{r}}(s_{t+1}) - V^i_{\phi_{r}}(s_{t})$. The value functions of role policy are parameterized as $\phi_{r}$. The state value is estimated by value functions $V^i_{r}$. Consequently, the optimization of the role policy gradient follows the equation:   
    
    \begin{equation}   
        \hat{g}^i_r =\hat{\mathbb{E}}_t[ \nabla_{\theta_{r}} \log \pi_{\theta_r}^{i}(\rho_{t}^{i} \mid o_{t}^{i},jo^i_t)\hat{A}^i_{r}]. 
    \end{equation}

    The main objective of our proposed role policy is represented as :

    \begin{equation}
        L_r = L_r^{CLIP} + c_1^{r} L_r^{KL} + c_2^{r} L_r^{VF} + c_3^{r} L_r^{E},
    \end{equation}

    where $L_r^{CLIP}$, $L_r^{KL}$, $L_r^{VF}$, and $L_r^{E}$ denotes the clipped surrogate objective, KL penalty, value function error of the role policy, and entropy bonus respectively; $c_1^{r}$, $c_2^{r}$, and $c_3^{r}$ represent coefficients for the KL penalty, value function and entropy, as introduced by schulman \emph{et al.} \cite{schulman2017proximal}. The surrogate loss is further degraded by the KL penalty $L_r^{KL}$ and $c_1^{r}$ is set to 0.5. The $c_2^{r}$ and $c_3^{r}$ are set as $1.0 \times 10 ^{-4}$ and 0.01, respectively.

\subsection{Learning the Conditional Primitive Policy}

    The training way of primitive policy is similar to the role policy, we adopt the MAPPO algorithm with the actor-critic structure. The decentralized actor generates the primitive actions $a_t$ used for interacting with the environment. These primitive actions are derived from the local observation, extracted from the state, and conditioned on the role action $\rho_t$. Moreover, the primitive actor or the policy is capable of exploring and covering. Upon receiving role action outputs from the role policy, the robot is required to execute corresponding sub-tasks. Specifically, the primitive policy focuses on the frontier cells when the role action \emph{explore} is sampled over the role probability distribution. The robot needs to cover more target cells if the role action \emph{cover} is sampled.

\subsubsection{Primitive Action Space}

    We utilize a 2D grid map to model the area search environment for multi-robot scenarios. The primitive action space comprises five discrete values: $\{Move\mbox{-}forward, Turn\mbox{-}right, Move\mbox{-}backward, Turn\mbox{-}left,\\ Stop\}$. These primitive actions are encoded as one-hot vectors and are determined based on the probability distribution output by the primitive policy Actor$^P$ or $\pi_{\theta_p}$.

\subsubsection{Primitive Observation Space}

     At timestep $t$, the local observation $o^i_t=\{o^{o}_t,o^{e}_t,o^{c}_t,o^{p}_t\}$ of robot $i$, detailed in Section V-A, which is encoded by an encoder $f(o^i_t): \mathcal{O} \to \mathbb{R}^F$. For our implementation, we adopt a Convolutional Neural Network (CNN) as the encoder, generating a feature vector $z_t^i$. The encoder module is shared among each robot. The role action $\rho_i^t$ of the robot $i$ and the feature vector of local observation $z_t^i$ are concatenated, forming the primitive observation with dimensions of $F+1$. 

    \begin{figure}[t]
        \centering
        \includegraphics[scale=0.4]{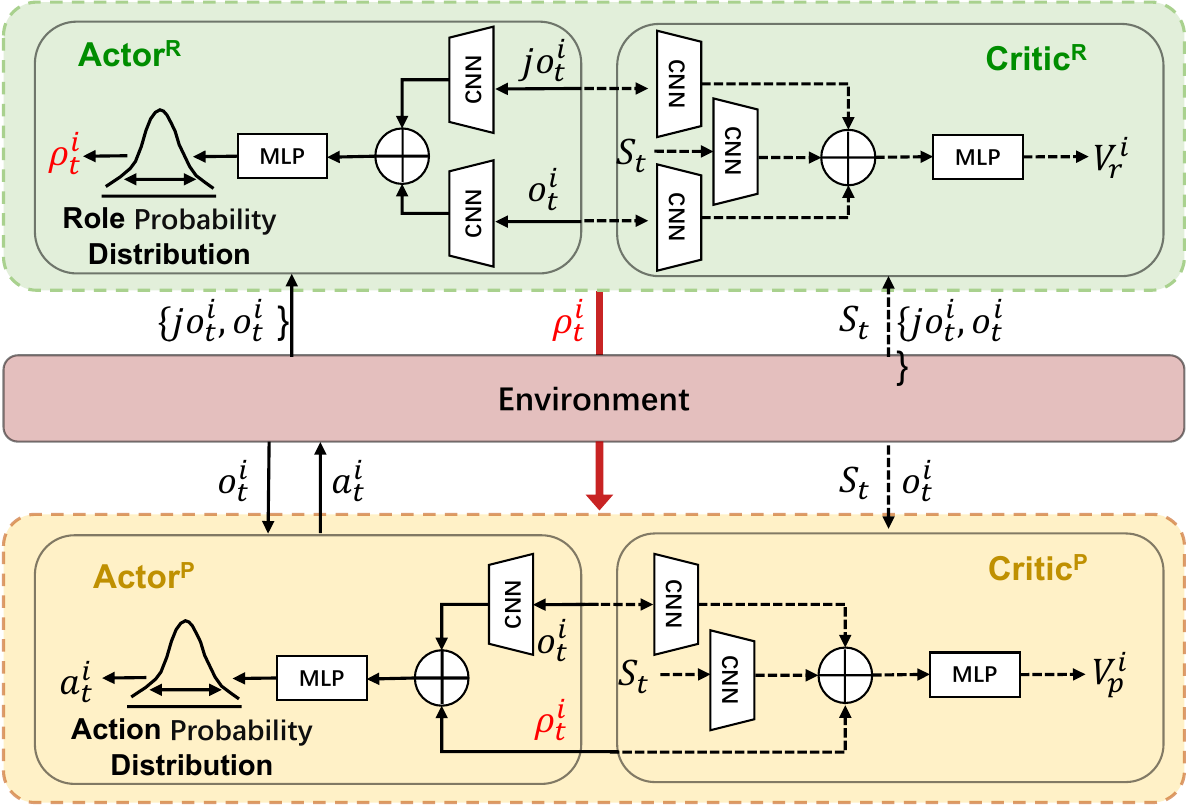}
         \caption{The training pipeline of our framework. We train the role and primitive policy by the MAPPO algorithm with two actor-critic structures. The green rectangle denotes the role-selection module and outputs the role action $\rho^i_t$, which is one of the inputs to the primitive policy. The actor policy is used for execution and training, which is based on local observation. The critic policy is only used for the training phases and is based on local observation and state information.}
        \label{fig:pipline}
        % \vspace{-0.5cm}
    \end{figure}
    \vspace{0.3cm}

\subsubsection{Primitive Rewards}

    For the training of primitive policy, we set two different rewards depending on the corresponding sub-tasks: exploration reward $R_e$ and coverage reward $R_c$. The primitive reward at timestep $t$ is denoted as $R_p (t)$. 

    % \begin{equation}
    %     \begin{aligned}
    %     R^p(t)=\left\{\begin{array}{ll}
    %     \sum\limits_{i \in N} R^e_i,\: R^e_i = \frac{\sum\limits_{k \in C_t} c_t^i \mathbb{I}_{q_t^i=p^k}}{B_e},  \;  \rho_{t}^i=0 \\
    %      \sum\limits_{i \in N} R^c_i, \: R^c_i = \sum\limits_{j \in C_t} R_t^i \mathbb{I}_{q_t^i=p^j} ,  \;  \rho_{t}^i=1
    %     \end{array}\right.
    %     \end{aligned}
    % \end{equation}

    \begin{equation}
     {R_p(t)} =
    \begin{cases}
          R_e,\: R_e = \sum\limits_{i \in N} \frac{\sum\limits_{k \in \bar{E_t}} u_t^k \mathbb{I}_{q_t^i=p^k}}{B_e}, & \text{if } \rho_{t}^i=0 \\
         R_c, \: R_c = \sum\limits_{i \in N} \sum\limits_{j \in C_t} c_t^j \mathbb{I}_{q_t^i=p^j}, & \text{if } \rho_{t}^i=1,
    \end{cases}
    \end{equation}

     where $\rho_{t}^i$ denotes the role action output by the Actor$^R$. In our fully cooperative multi-robot setting, the robots with the same role are trained using a shared global reward. The role reward at timestep $t$ is the sum of the local reward with the same role. As for setting the coverage reward, when a robot visits a target cell $p^j$ ($q_t^i=p^j$), it receives a coverage reward of 1. The exploration radius of each robot is defined as $rad_e$, allowing the robot to explore up to $2\pi \cdot rad_e$ cells. During training, we accumulate the number of unexplored cells $u^k$ ($q_t^i=p^k$) within the exploration range $\sum\limits_{k \in C_t} u_t^k \mathbb{I}_{q_t^i=p^k}$ at the timestep $t$. To normalize the exploration reward within the range of $(0, 1)$ and align it with the coverage reward, the exploration area is divided by an exploratory ability $B_e$: 

    \begin{equation}
    \begin{gathered}
    B_e = 2 \pi \cdot rad_e^2 - S_{inter} \\
    S_{inter} = 2rad_e^2 \ast arccos(\frac{d^2}{2d \cdot rad_e})- \sqrt{4rad_e^2 -d^2},
    \end{gathered}
    \end{equation}

    where $S_{inter}$ denotes the intersection of explored areas at different two timesteps, and $d$ denotes the distance between these two timesteps. Considering the discrete primitive action space, where the robot can move only one grid, we set $d$ as 1. Furthermore, $B_e$ denotes the maximum exploration area.  

\subsubsection{Primitive Policy Optimization}

    In the primitive policy, each robot learns a local primitive policy $\pi_{\theta_p}^{i}\left(a_{t}^{i} \mid o_{t}^{i},\rho_{t}^{i}\right)$ and parameterized by a local set $\theta_p$. The GAE function of the primitive policy, $\hat{A}_p^i = \sum_{t=1}^{\infty} (\gamma\lambda)^l \delta^i_p (t+l)$ and TD-error is $\delta^i_p (t) = r^i_t + \lambda V^i_{\phi_{p}}(s_{t+1}) - V^i_{\phi_{p}}(s_{t})$. 
    The state value of agents under the primitive policy is estimated by functions $V^i_{p}$ parameterized by $\phi_{p}$. Therefore, the processing of primitive policy gradient ascent follows the equation:   
    
    \begin{equation}   
        \hat{g}^i_p =\hat{\mathbb{E}}_t[ \nabla_{\theta_{p}} \log \pi_{\theta_p}^{i}(a_{t}^{i} \mid o_{t}^{i},\rho_{t}^{i})\hat{A}^i_{p}]. 
    \end{equation}

    Subsequently, the primitive actions are sampled from an action probability distribution produced by the primitive policy, i.e. $a^{i}_{t} \sim \pi_{\theta_p}^{i}\left(a_{t}^{i} \mid o_{t}^{i},\rho_{t}^{i}\right)$.

    The main objective of the proposed primitive policy is represented as :

    \begin{equation}
        L_p = L_p^{CLIP} + c_1^{p} L_p^{KL} + c_2^{p} L_p^{VF} + c_3^{p} L_p^{E},
    \end{equation}

    where $L_p^{CLIP}$, $L_p^{KL}$, $L_p^{VF}$, and $L_p^{E}$ are the clipped surrogate function, KL penalty, value function error of the primitive policy, and entropy bones respectively; $c_1^{p}$, $c_2^{p}$, and $c_3^{p}$ are coefficients of KL penalty, value function and entropy.

\subsection{Overall Optimization Objective}

    We establish optimization objectives for learning role and primitive policy. The decentralized policy of each robot is shared and updated via the gradients derived from the corresponding standard Temporal Difference (TD) loss in reinforcement learning. Networks are trained using data collected from the trajectory of all robots, sampled from the replay buffer. Hence, our final learning objective is defined as:
    
    \begin{equation}
        L = L_p + L_r,
    \end{equation}

    where $L_p$ and $L_r$ are the main objectives of the role and primitive policy, respectively. Theoretically, optimizing the role policy $\pi_{\theta_{r}}$ and primitive policy $\pi_{\theta_{p}}$ alternatively with different samples from the replay buffer is advisable. In this work, the gathered data conforming to the GA-POMDP contain information on both role and primitive policy. In practice, optimizing $\pi_{\theta_{r}}$ and $\pi_{\theta_{p}}$ using the same samples significantly enhances data efficiency. We adopt the framework of centralized training and decentralized execution structure, deploying only the role actor and the primitive actor networks during execution.

%===============================================================================

\section{Experiments}

    In this section, we conduct experiments to answer the following questions: (1) How does the performance of our method compare with the existing research? (2) Can our method be extended to more complex environments? (3) Can our method be applied to scenarios involving a larger number of robots? (4) Why is the role-selection framework practical? The well-designed experiments validate the effectiveness of our method and its potential application in realistic scenarios.

\subsection{Experiment Setup}

\begin{figure}[t]
        \vspace{0.5em}
        \centering
        \includegraphics[scale=0.3]{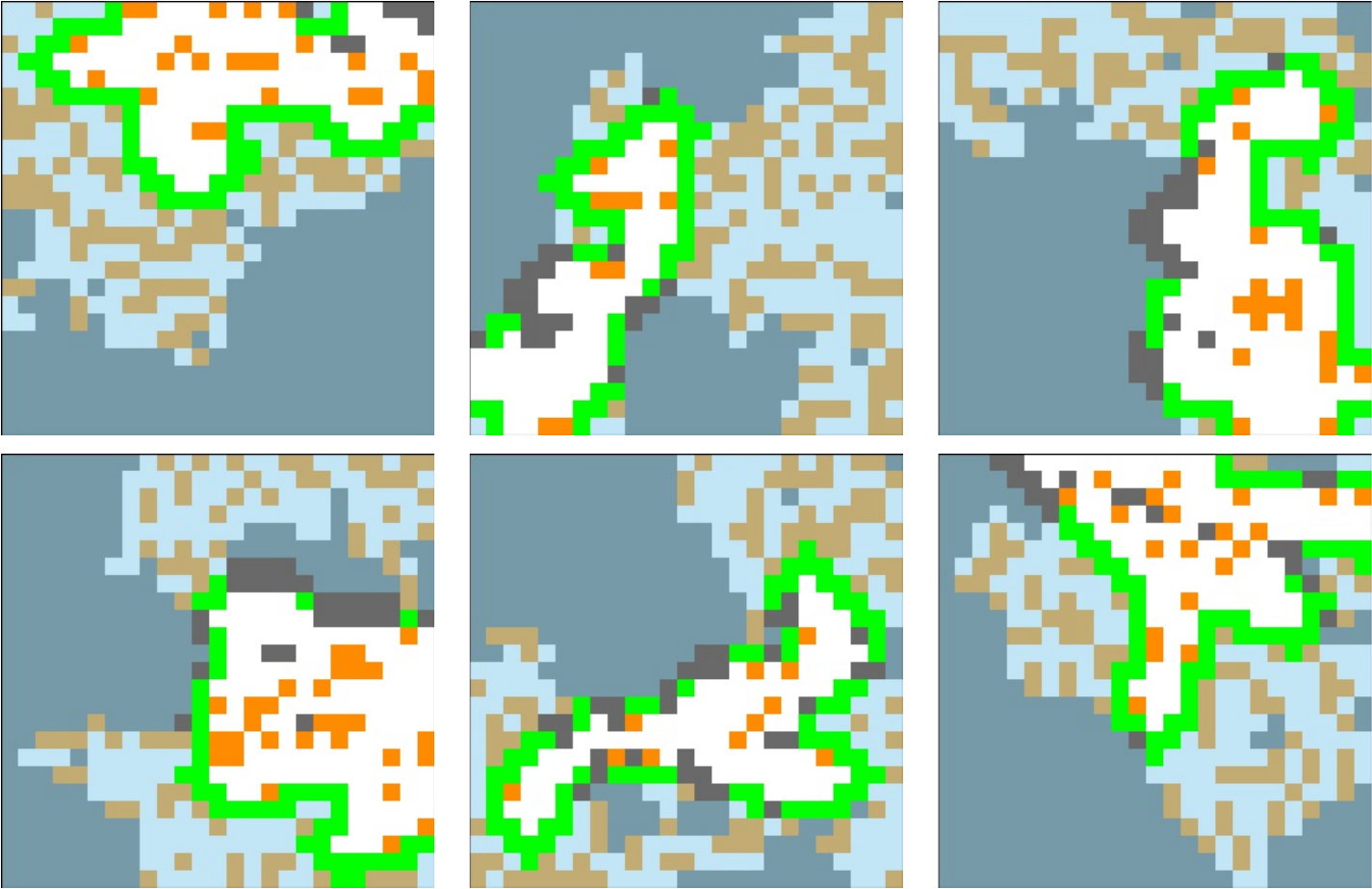}
        \caption{Map examples. The train and test scenes are randomly generated by the gym simulation environment. The grey, orange, and green cells represent the obstacle, target, and frontier cells respectively. Besides, The obstacles are randomly generated while maintaining the connectivity of the environment.}
        \label{fig:simulation_envs}
        \vspace{-0.3cm}
    \end{figure}

    We train and test our method in the randomly generated gym simulation environment, as shown in Fig. \ref{fig:simulation_envs}. The size of simulated maps is $25 \times 25$, with 250 obstacles and 100 target cells. In the subsequent well-designed experiments, we deploy more robots in the more complex simulated maps, altering the number of targets or obstacles. The field of the view is set to 4, and the communication radius is set to 10. 

    \textbf{Network Architecture:} Before adopting a local Multi-layer Perception (MLP) to train role or primitive actor network, we construct the Convolutional Neural Networks (CNN) and MLP to encode the local observation. A similar structure is designed to train the role or primitive critic network. 
    
    The training and execution processing is illustrated in Fig. \ref{fig:pipline}. For the training of role actor network, we encode the joint local observation $jo^i_t$ using a CNN architecture, constructed by sequentially applying $Conv2d\mbox{-}BatchNorm2d\mbox{-}ReLU\mbox{-}MaxPool2d$ and $Conv2d\mbox{-}BatchNorm2d\mbox{-}ReLU$ blocks twice. The local observation $o_t^i$ is encoded using a 3-layer CNN architecture, which then serves as input to a 3-layer MLP with hidden layers of 128, 64, and 32 units respectively, generating the feature vectors of local observation. These feature vectors of local observation are further input into a GNN architecture to generate an aggregated feature vector of 128 dimensions. The aggregated feature vector is then fed into a 2-layer MLP with hidden layers of 64 and 32 units respectively, producing the role distribution, Finally, the role action $\rho^i_t$ is sampled from this role distribution. For the training of the primitive actor network, we stack the feature vector of local observation and role action and then input it into a 3-layer MLP.
    
    \textbf{Training Parameter Settings:} Our model was trained using an Intel(R) Core(TM) i9-12900K CPU and an Nvidia GeForce RTX 4080 GPU in Pytorch, taking approximately 12 hours to complete $5\times10^6$ timesteps. We employ distribution training through the Ray \cite{moritz2018ray} and Rllib \cite{liang2018rllib}. In all training algorithm configurations, the optimization is conducted with the learning rate of $5\times 10^{-4}$, the PPO clipping parameter $\varepsilon\mbox{-}greedy$ is set to 0.2, and the discount factor is set to 0.9. The training batch and minibatch sizes are set to 5000 and 1000, respectively. 

    \textbf{Evaluation Metrics:} We evaluate the performance of our method in comparison with the state-of-the-art methods in terms of percentage and efficiency of task completion. The percentage of task completion is related to the sub-tasks of multi-robot area search, encompassing two parts: exploration and coverage percentage. The efficiency in executing these sub-tasks demonstrates the collaboration among robots. Below are the specific evaluation metrics:

    1) \emph{Exploration percentage (Explo ($\%$))}: At each timestep, the explored area of the environment equals the union of the sensor ranges of all robots. The total explored area per episode sums the explored area from the initial test timestep to the maximum test timestep. The exploration percentage is calculated by averaging the explored area over 500 test episodes, divided by the entire free area excluding obstacle cells.

    2) \emph{Coverage percentage (Cover ($\%$))}: This metric represents the cumulative covered targets divided by all coverable targets. Only the area that has been explored allows the robot to locate the targets randomly generated within it. Thus, this metric heavily relies on the exploration percentage.
    
    3) \emph{Exploration efficiency (Time$\_$E (step))}: This metric measures the average consumed timesteps required to explore 90\% of the entire unknown environment.

\subsection{Baseline Methods}

    We verified the superiority of our method through a comprehensive comparison with various baseline methods in the well-designed experiments. The baseline methods encompass three distinct types: heuristic methods, expert methods, and state-of-the-art (SOTA) methods. Specifically: 

    \textbf{Random:} This algorithm involves each robot randomly choosing an action from the primitive action space for interaction with the environment.

    \textbf{Greedy:} In this approach, each robot selects the nearest cell from the frontier or target cells as its sub-goal. After reaching the sub-goal, the robot calculates the shortest Manhattan distance between its current location and the frontier or target cells to determine the subsequent sub-goal for the next timestep.

    \textbf{VRPC:} The VRP-based controller (VRPC) is an expert controller method detailed in \cite{tolstaya2021multi}, employing an optimization-based expert solution generated via Google's OR-Tools library \cite{c53}.

    \textbf{GNN:} This algorithm accomplishes the simultaneous execution of two sub-tasks by learning the spatial relationship on a graph-represented environment, which is presented in \cite{tolstaya2021multi}.

    \textbf{H2GNN:} Proposed in \cite{zhang2022h2gnn}, this algorithm applies a multi-head attention mechanism within the graph neural network to differentiate the importance between exploration and coverage.

    \textbf{ECC:} This is a variant end-to-end method directly from the local observation to decision-making and proposed in \cite{blumenkamp2021emergence}. We applied this method to multi-robot area search, and the efficiency of the role-selection framework is verified by comparing ECC and our method in terms of the third metric.

\subsection{The Performance Analysis}
    
    The results in Table \ref{table:4robots}, \ref{table:8robots}, and \ref{table:15robots} prove the superior performance of our method in terms of generalization and scalability within the well-designed scenes. We validate the generalization by deploying varying robots into scenes, specifically 4, 8, and 15 robots. Meanwhile, we assess the scalability of our method on the four scenes, comprising easy, medium, hard, and super hard scenes, which is determined by the proportions of randomly generated target cells: $56\%$, $24\%$, $16\%$, and $8\%$ respectively. A higher ratio of target cells in a scene facilitates recognition and coverage of the targets by the robots.

    \begin{table*}[]
        \renewcommand{\arraystretch}{1}
        \centering
        \caption{Quantitative Results of Deploying 4 Robots in Different Environments}
        \label{table:4robots}
        \begin{tabular}{c|c c| c c|c c| c c}
            \toprule
            & \multicolumn{2}{c|}{Easy Scene } & \multicolumn{2}{c|}{Medium Scene}& \multicolumn{2}{c|}{Hard Scene} & \multicolumn{2}{c}{Super hard Scene }\\
            \cmidrule{0-1}\cmidrule{2-3} \cmidrule{4-5} \cmidrule{6-7} \cmidrule{8-9} 
            Method & Explo (\%) & Cover (\%) & Explo (\%) & Cover (\%)& Explo (\%) & Cover (\%) & Explo (\%) & Cover (\%) \\
            \hline
            Random & 22.1 & 11.4 & 22.2 & 11.3& 22.2 & 11.2 & 22.2 & 11.0\\
            Greedy & 70.2  & 34.6 & 75.2&38.1&77.0 &39.1  & 78.4  & 40.2\\
            VRPC & 60.8 & 41.6 & 72.6 & 53.9 & 72.9 & 63.6 & 67.7 & 59.6\\
        % \hline
            GNN  & 82.9& 35.1 & 82.1 & 35.6 & 82.4 &36.4 &   79.7 &35.6 \\
            H2GNN & 79.8 & 43.6 & 81.8 & 54.8& 81.7& 57.1 &   83.4 & 63.7 \\
        % \hline
            ECC & 79.7 & \textbf{46.8} & 83.3 & \textbf{59.9} & 81.8 & 62.9 & 79.1 & 64.1\\
            Ours & \textbf{99.5} & 45.2 & \textbf{97.9}& 59.3& \textbf{98.9} & \textbf{64.1} & \textbf{100}& \textbf{69.5} \\
            \bottomrule
        \end{tabular}
    \end{table*}

    \begin{table*}[]
        \renewcommand{\arraystretch}{1}
        \centering
        \caption{Quantitative Results of Deploying 8 Robots in Different Environments}
        \label{table:8robots}
        \begin{tabular}{c|c c| c c|c c|c c}
            \toprule
             & \multicolumn{2}{c|}{Easy Scene } & \multicolumn{2}{c|}{Medium Scene}& \multicolumn{2}{c|}{Hard Scene} & \multicolumn{2}{c}{Super hard Scene }\\
            \cmidrule{0-1}\cmidrule{2-3} \cmidrule{4-5} \cmidrule{6-7} \cmidrule{8-9} 
            Method & Explo (\%) & Cover (\%) & Explo (\%) & Cover (\%)& Explo (\%) & Cover (\%) & Explo (\%) & Cover (\%) \\
            \hline
            Random & 34.7 & 20.5& 34.7 & 20.4 & 24.7 & 20.2 & 24.7 & 20.0\\
            Greedy & 82.1 & 47.1 & 89.5& 52.3 & 91.3 & 53.8 & 93.1 & 55.0\\
            VRPC & $-$ &$ -$& $-$& $-$ &$ -$ & $-$ & $-$ & $-$\\
            % \hline
            GNN  & 87.7 & 53.0& 87.5& 53.5& 82.0& 50.3 & 84.6& 50.9\\
            H2GNN & 86.5& 65.1 & 85.1 & 69.9 & 85.1 & 71.8 & 85.5 & 74.3 \\
            % \hline
            ECC & \textbf{100}& 62.1 &  \textbf{100} & 79.0& \textbf{100}& 81.3 & \textbf{100} & 81.7\\
            Ours & \textbf{100}&\textbf{69.3}& \textbf{100} & \textbf{79.7}& \textbf{100}& \textbf{82.9}& \textbf{100}& \textbf{85.4}\\
            \bottomrule
            \multicolumn{9}{l}{\makecell[l]{$-$ \small{denotes the VRPC doesn't function in large-scale scenes with more robots due to its NP-hard nature},\\
            \small{resulting in computational local surpassing the capacity of computers as the number of robots increases.}}}\\
        \end{tabular}
    \end{table*}

    \begin{table*}[]
        \renewcommand{\arraystretch}{1}
        \centering
        \caption{Quantitative Results of Deploying 15 Robots in Different Environments}
        \label{table:15robots}
        \begin{tabular}{c|c c |c c|c c| c c}
            \toprule
            & \multicolumn{2}{c|}{Easy Scene } & \multicolumn{2}{c|}{Medium Scene}& \multicolumn{2}{c|}{Hard Scene} & \multicolumn{2}{c}{Super hard Scene }\\
            \cmidrule{0-1}\cmidrule{2-3} \cmidrule{4-5} \cmidrule{6-7} \cmidrule{8-9} 
            Method & Explo (\%) & Cover (\%) & Explo (\%) & Cover (\%)& Explo (\%) & Cover (\%) & Explo (\%) & Cover (\%) \\
            \hline
            Random & 26.1 & 20.5& 26.1 &20.4 & 26.1 & 20.4& 26.1 & 20.3\\
            Greedy & 92.7 & 56.9 & 100 & 63.1&  100 & 64.5& 100 & 65.5\\
           VRPC & $-$ &$ -$& $-$& $-$ &$ -$ & $-$ & $-$ & $-$\\
            % \hline
            GNN  & 89.1& 66.9&88.1 & 66.4& 87.4 & 65.0 & 89.0 & 65.3  \\
            H2GNN & 88.7 & 77.4 &  89.0 & 77.6  & 88.1 & 77.4 & 86.4  & 76.2\\
            % \hline
            ECC & \textbf{100} & \textbf{85.3}& \textbf{100} & 87.5& \textbf{100} & 88.1 & \textbf{100} & 88.3\\
            Ours & \textbf{100}& 84.0& \textbf{100}& \textbf{88.5} & \textbf{100} & \textbf{89.8} &\textbf{100}& \textbf{91.5}\\
            \bottomrule
        \end{tabular}
    \end{table*}

    \begin{table*}[t]
        \renewcommand{\arraystretch}{1}
        \centering
        \caption{Quantitative Results of Environments with Different Numbers of Obstacles}
        \tabcolsep=0.41cm  % 设置表格间距
        \label{table:3obstacles}
        \begin{tabular}{c|c c|c c|c c}
            \toprule
            & \multicolumn{2}{c|}{Easy Scene (coverable 0.84)} &
            \multicolumn{2}{c|}{Medium Scene (coverable 0.6)}& 
            \multicolumn{2}{c}{Hard Scene (coverable 0.49)}\\
            \cmidrule{0-1}\cmidrule{2-3} \cmidrule{4-5} \cmidrule{6-7} 
            Method & Explo (\%) & Cover (\%)& Explo (\%) & Cover (\%)& Explo (\%) & Cover (\%)\\
            \hline
            Random & 18.8 & 9.6 & 22.2 & 11.2 & 24.9 & 12.6\\
            Greedy & 60.2& 31.1 & 77.0 & 39.1 & 89.6& 45.6 \\
            VRPC & 84.8& 70.6 &72.9 & 63.6 & 54.1 & 49.0\\
            % \hline
            GNN  & \textbf{97.8}& 34.5& 82.4& 36.4 & 69.4 & 34.2  \\
            H2GNN &97.5 & 61.1 & 81.7 & 57.1 & 70.2& 52.7  \\
            % \hline
            ECC & 69.4& 59.7 & 81.8& 62.9 & 91.8& 66.0 \\
            Ours & 89.1 & \textbf{61.5}& \textbf{98.9}& \textbf{64.2} & \textbf{100} & \textbf{66.2}\\
            \bottomrule
            % \multicolumn{9}{l}{\small $-$ {denotes the maximum test timesteps, while the percentage is less than 90\%.}}\\
        \end{tabular}
    \end{table*}

     Answering the first and second questions, as a horizontal comparison of the Tables \ref{table:4robots}, \ref{table:8robots}, and \ref{table:15robots}, the generalization of our approach is validated, which demonstrates superior performance in achieving the highest percentage of coverage and exploration. Notably, VRPC cannot function in environments with more than 8 robots due to its NP-hard nature, causing computational load beyond the capacity of the computer with the growing number of robots. Our method exhibits minimal advantages compared to baseline methods in the easy scene. Because in the easy scene with more frontier and target cells, the robots can find more. As scene complexity increases, the performance of Random, GNN, and H2GNN displays minimal fluctuation. However, these three methods exhibit limited adaptability in a high-complexity environment, indicating poor generalization. Conversely, Greedy, ECC, and our approach showcase a higher coverage percentage in the super hard scene compared to other scenes. This result stems from the robots effectively locating and covering fewer target cells in the challenging super hard scene. This substantiates our approach's competence in selecting targets or frontier cells with the least global planning cost, akin to the Greedy method, facilitating efficient exploration within shorter test timesteps.

     \begin{table*}[]
        \renewcommand{\arraystretch}{1}
        \centering
        \caption{Ablation Studies the efficacy of our method across four scenes with more robots}
        \label{table:ablation_study}
        \begin{tabular}{c|c|c c| c c|c c| c c}
            \toprule
                & & \multicolumn{2}{c|}{Easy Scene } & \multicolumn{2}{c|}{Medium Scene}& \multicolumn{2}{c|}{Hard Scene} & \multicolumn{2}{c}{Super hard Scene }\\
                \cmidrule{0-1}\cmidrule{1-2}\cmidrule{3-4} \cmidrule{5-6} \cmidrule{7-8} \cmidrule{9-10} 
                Method& N  & Explo (\%) & Time\_E (step) & Explo (\%) & Time\_E (step)& Explo (\%) &Time\_E (step) & Explo (\%) & Time\_E (step) \\
            \hline
                ECC& \multirow{2}{*}{4}  & 79.5 & $-$ & 83.3 & $-$ & 81.8 & $-$  & 79.1 & $-$ \\
                Ours&   & \textbf{99.5}  & \textbf{39} & \textbf{97.9}&\textbf{42} &\textbf{98.9} &\textbf{41}  & \textbf{100}  & \textbf{39}\\
            \hline
               ECC& \multirow{2}{*}{8}  & 100 & 38 & 100 & 37 & 100 & 37  & 100 & 38 \\
                Ours&  & 100  &\textbf{28} & 100&\textbf{30} &100 &\textbf{29}  & 100  & \textbf{27}\\
            \hline
                ECC& \multirow{2}{*}{15}  & 100 & 29 & 100 & 29 & 100 & 29  & 100 & 28 \\
                Ours&   & 100  & \textbf{23} & 100&\textbf{25} &100 &\textbf{24}  & 100  & \textbf{23}\\
            \bottomrule
            \multicolumn{9}{l}{\small $-$ {denotes the maximum test timesteps, while the percentage is less than 90\%, N denotes the number of the robots.}}\\
        \end{tabular}
    \end{table*}

    \begin{figure*}[htb]
        \centering
        \subfloat[]{\includegraphics[scale=0.45]{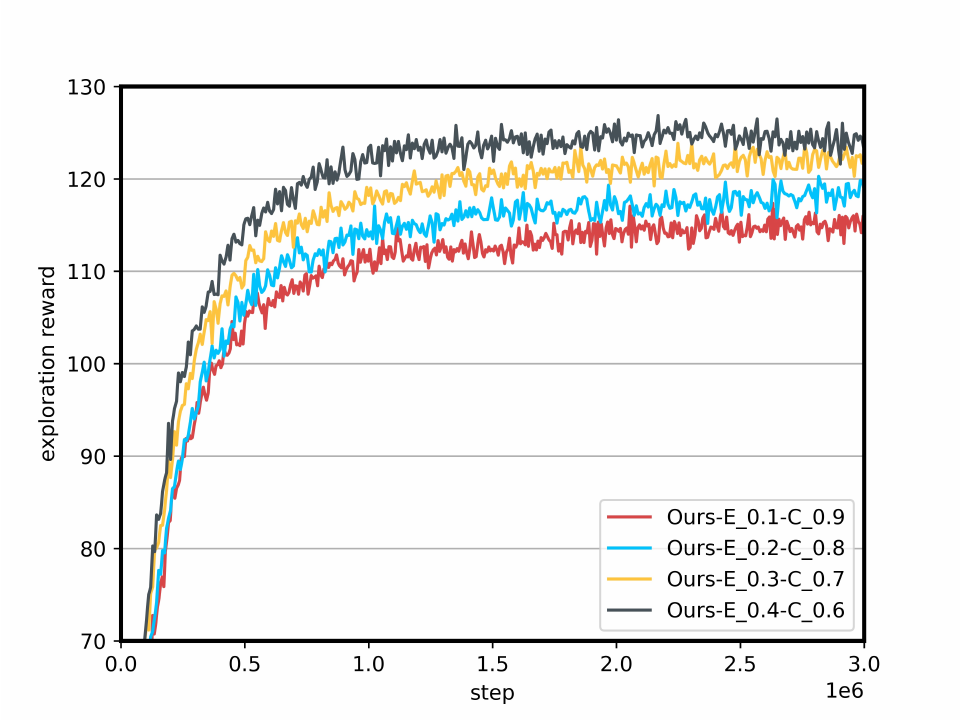}
        \label{fig:train_all_role_ratio_in_exploration}}
        \hfil
        \subfloat[]{\includegraphics[scale=0.45]{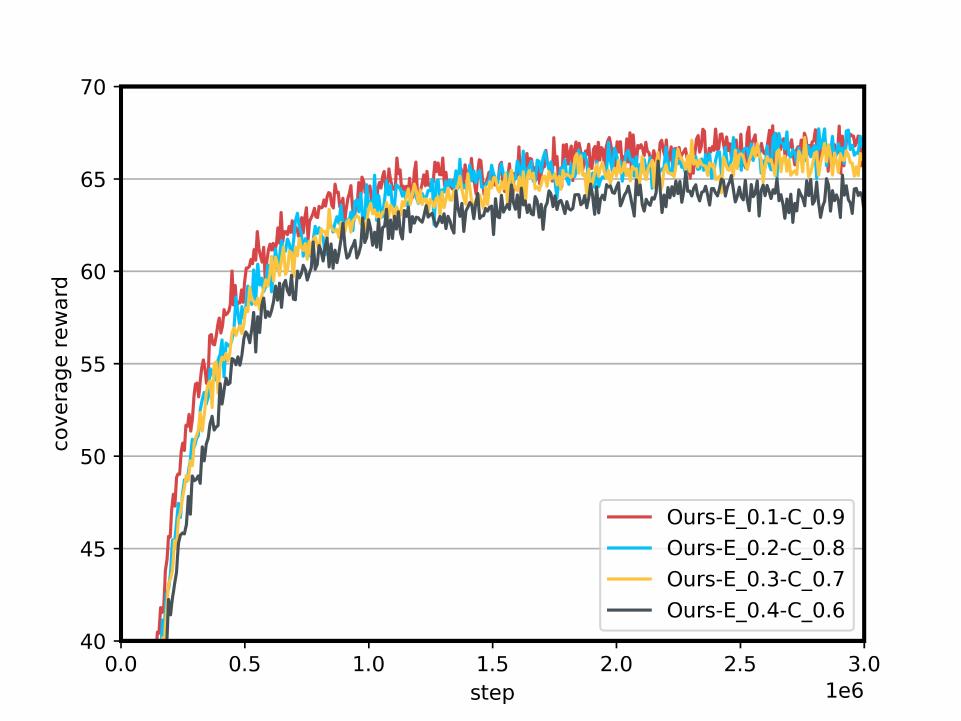}
        \label{fig:train_all_role_ratio_in_coverage}} 
        \caption{Performance comparison of the training phases with different weigh coefficients of exploration and coverage roles, compared against baseline methods. `E' and `C' represent the exploration and coverage, respectively. The values less than 1 alongside the roles indicate the reward weights coefficients assigned to respective sub-tasks, with the combined weights of `E' and `C' adding up to 1. (a) Comparative performance regarding exploration rewards. (b) Comparative performance regarding coverage rewards.}
        \label{fig:train_resulting}
    \end{figure*}

    \begin{figure*}[htb]
        \centering
         \subfloat[]{\includegraphics[scale=0.45]{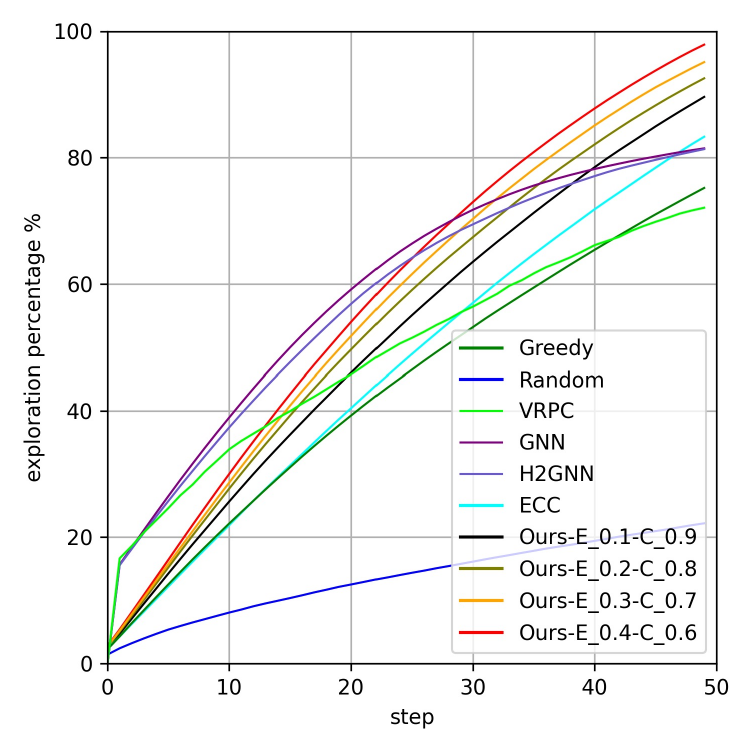}
        \label{fig:test_in_exploration}}
        \hfil
        \subfloat[]{\includegraphics[scale=0.45]{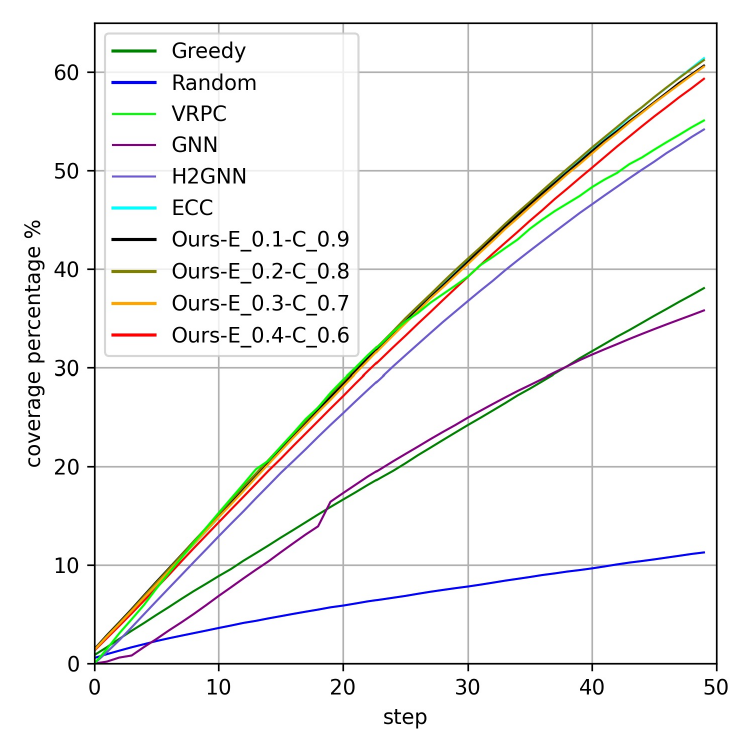}
        \label{fig:test_in_coverage}} 
        \caption{Performance of the testing scenes with varying role weight coefficients for exploration and coverage, compared with baseline methods. (a) Exploration percentage. (b) Coverage percentage. Both metrics validate the effectiveness of sub-tasks: the exploration percentage demonstrates the exploration capability (shown in (a)), while the coverage percentage illustrates the ability in coverage (shown in (b)).}
        \label{fig:test_resulting}

    \end{figure*}

    \begin{figure*}[htb]
        \centering
        \subfloat[]{\includegraphics[scale=0.45]{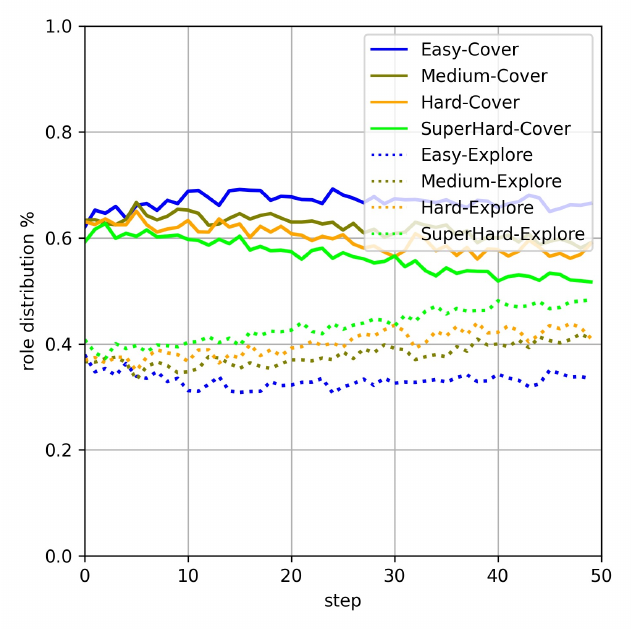}
        \label{fig:role_distribution_4v6}} 
        \hfil
        \subfloat[]{\includegraphics[scale=0.45]{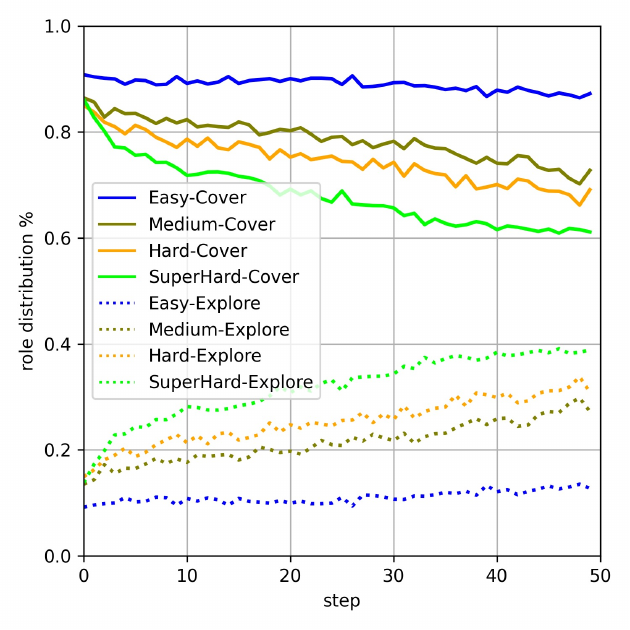}
        \label{fig:role_distribution_3v7}} 
        \hfil
         \subfloat[]{\includegraphics[scale=0.45]{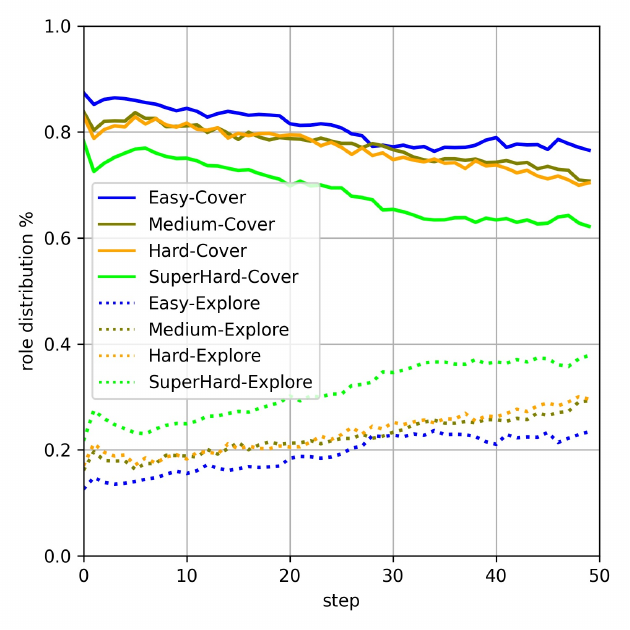}
        \label{fig:role_distribution_2v8}}
        \caption{The experimental results that analyze roles output proportions across four scenes, easy, medium, hard, and super hard scenes. This figure depicts the output role proportions under different role weight coefficient settings: (a) Exploration at 0.4 and coverage at 0.6. (b) Exploration at 0.3 and coverage at 0.7. (c) Exploration at 0.2 and coverage at 0.8. Solid lines denote \emph{cover} role proportions, while dashed lines represent \emph{explore} role proportions.}
        \label{fig:role_distribution} 
    \end{figure*}

    Answering the first and third questions, through a vertical comparison of the Tables \ref{table:4robots}, \ref{table:8robots}, and \ref{table:15robots}, our approach demonstrates superior scalability in terms of exploration and coverage percentage. Despite deploying more robots, the Random, GNN, and H2GNN methods do not significantly enhance their performance. Random methods performed poorly due to the robot's constrained planning abilities. Similarly, the inferior performance of the GNN and H2GNN methods is attributed to the robot's inability to disperse effectively for exploration and coverage, a crucial factor for cooperative efforts among the robots. As the number of robots increased, Greedy, ECC, and ours notably improved, particularly in achieving a higher coverage percentage (all these methods reached $100\%$ exploration percentage). This highlights the robots' ability to intelligent switch between exploration and coverage, prioritizing coverage within sufficient explored areas. This adaptation supports the robust scalability of our approach.

    To validate the robust long-term planning ability of our method, we conducted the test using all baseline methods in three scenes employing 4 robots. These scenes are characterized by varying obstacle ratios of $16\%$, $40\%$, and $52\%$, corresponding to the easy, medium, and hard scenes, respectively (ref Table \ref{table:3obstacles}). In the easy scene with fewer obstacles, the ECC and our work exhibited inferior performance in terms of the exploration percentage compared to GNN and H2GNN. This can be attributed to the superior local planning capability, facilitating rapid identification and coverage of frontier and target cells during exploration. However, as the environment complexity escalated with more obstacles, Greedy, ECC, and our work outperformed the GNN and H2GNN in terms of coverage percentage. This improved performance is due to their strong capability in obstacle avoidance and focus on the frontier and target cells within the global field of view, which further verifies the robust capability in long-term planning. Furthermore, both ECC and our work showcase the higher exploration and coverage percentage compared to the Greedy, indicating that the robots learn to disperse planning, thereby reducing the duplicate exploration or coverage. 

\subsection{The Efficiency Analysis}
    Although the higher percentage of coverage and exploration demonstrates the better advantage of our method, the efficiency of higher performance is considered to achieve faster area searching. We develop an analysis of our method's efficacy in task completion, as detailed in Table \ref{table:ablation_study}. We compared the performance of ECC with our work in terms of the exploration percentage and consumed timestep to reach $90\%$ exploration percentage. We conduct experiments with four scenes with varying robot numbers, including 4, 8, and 15. The well-designed studies indicate that our method outperforms ECC in terms of both the exploration percentage and consumed timestep in the scene involving 4 robots. As the number of robots increases, the exploration percentage of both two methods becomes comparable, eventually achieving complete exploration. Furthermore, our method demonstrates a lower consumed timestep to reach the $90\%$ exploration percentage compared to ECC. This signifies the robust cooperative exploration capability among the robots, and showcases their ability to effectively disperse planning and minimize duplicate exploration.

\subsection{Analysis of Role Weight Influence}

    We conduct additional experiments with varying role weights to further evaluate the performance of the role-selection module, and answer the last question. We more focus on the coverage of the area search tasks, so we set a higher weight coefficient for coverage than for exploration. Specifically, the coverage weight coefficient is set at 0.6, 0.7, 0.8, and 0.9, while the corresponding exploration coefficients are 0.4, 0.3, 0.2, and 0.1. Thus, we compared the performance of our method in the medium scene with four role coefficients, illustrating training results in Fig. \ref{fig:train_resulting}\subref{fig:train_all_role_ratio_in_exploration} and \ref{fig:train_resulting}\subref{fig:train_all_role_ratio_in_coverage} and in terms of exploration and coverage reward. From the results, increasing the exploration weight augments the exploration reward, while a larger coverage factor correlates with a higher coverage reward. The performance of all methods, including our work with different four coefficients of role in testing scenes are shown in Fig. \ref{fig:test_resulting}\subref{fig:test_in_coverage} and \ref{fig:test_resulting}\subref{fig:test_in_exploration} in terms of exploration and coverage percentage. Within the maximum test timestep, the exploration percentage of our method with different coefficients significantly outperforms other methods, including the heuristic and SOTA approaches.

    To further analyze the role proportions in the testing scenes with various role weight coefficients in our work. We conducted tests to analyze the proportions of role outputs in the easy, medium, and super-hard scenes, as illustrated in Fig. \ref{fig:role_distribution}. In our method, with exploration and coverage weight coefficients set as 0.4 and 0.6 respectively, the \emph{cover} role proportions outputs by the role policy increase as the number of target cells grows. In the scene where the number of target cells surpasses the frontier cells, the robots are inclined towards covering rather than exploring. Consequently, the \emph{explore} role proportions remain at its lowest across all scenes. As the coverage weight coefficient increases but stays below 0.7, the \emph{cover} role proportions rise. This substantiates the effectiveness of maintaining a balance between exploration and coverage by adjusting the reward allocation for respective sub-tasks. However, in the scene where the coverage weight coefficient exceeds 0.7, Fig. \ref {fig:role_distribution}\subref{fig:role_distribution_2v8} illustrates an increase in \emph{explore} role proportions, resulting in a decrease in \emph{cover} role proportions. This shift occurs due to the initial stages of task execution, marked by a notable overlap between frontier and target cells, leading to a higher number of cells along the coverage direction. Consequently, the network prioritizes exploration in the early initial stages, displaying a greater inclination toward this aspect.

%===============================================================================

\section{Conclusion and Discussion}

    The concept of role provides the capability for robots to understand and learn task planning from the upper-view for the multi-robot area search. In this paper, we propose a role-selection algorithm comprising a role policy and a primitive policy. The role policy learns a mapping from the local and joint observation to the role actions, which is crucial for robots to learn who they are. The role-switching enables the role-selection module to function between two timesteps. The primitive policy learns ``how" to plan conditioning on the role action outputs by the role policy. The primitive policy possesses more skills, which means that regardless of the role, it could learn to execute the corresponding decisions. Moreover, to further promote cooperation among the robots, both role and primitive policy are trained using a multi-agent reinforcement learning algorithm with a double actor-critic structure. We design the seven scenes with varying complexities to verify the scalability and generalization of our method. Furthermore, we also verify the effectiveness of our work in terms of the consumed timesteps of the explored area reaching 90\%. Finally, we analyze the impact of role weight on the performance of exploration and coverage.
    
    While our method demonstrates superiority in terms of exploration and coverage percentage, consumed timestep over $90\%$ exploration area, it does not achieve complete coverage. The improvement of the coverage percentage is evident by adjusting the weight coefficient of coverage, which is heuristic and designed manually. As a result, our focus will shift to researching intelligent methods for dynamically adjusting the role weight coefficient based on environment states. Besides, the representation of the frontier cells, target cells, and obstacle cells in a binary form, followed by encoding into embedding vectors using CNN, may result in weakened environmental representation, particularly in complex scenes. To enhance this, we plan to enrich visual perception by implementing algorithms into the 3D environment \cite{wasserman2023last}.

%===============================================================================

% \section*{Acknowledgments}
% This should be a simple paragraph before the References to thank those individuals and institutions who have supported your work on this article.

\bibliographystyle{IEEEtran}
\bibliography{root.bib}

% \end{IEEEbiography}

\end{document}